\pdfoutput=1

\documentclass[11pt]{article}

\usepackage[]{acl}

\usepackage{times}
\usepackage{latexsym}

\usepackage[T1]{fontenc}

\usepackage[utf8]{inputenc}

\usepackage{microtype}

\usepackage{adjustbox}
\usepackage{array}
\usepackage{bm}
\usepackage{booktabs}
\usepackage{multirow}
\usepackage{subfigure}
\usepackage{color}
\usepackage{mathtools}
\usepackage{paralist}

\usepackage{algorithm}
\usepackage{algorithmicx}
\usepackage{algpseudocode}
\usepackage{amsfonts,amssymb} 
\usepackage{amsmath}
\usepackage{hyperref}

\title{Implicit Relation Linking for Question Answering over Knowledge Graph}

\author{
    Yao Zhao$^{1}$, \ 
    Jiacheng Huang$^{1}$, \ 
    Wei Hu$^{1,2,}$\thanks{~~Wei Hu is the corresponding author.}, \\
    \textbf{Qijin Chen}$^{3}$, \
    \textbf{Xiaoxia Qiu}$^{3}$, \
    \textbf{Chengfu Huo}$^{3}$, \
    \textbf{Weijun Ren}$^{3}$ \\ 
    $^{1}$State Key Laboratory for Novel Software Technology, Nanjing University, China \\
    $^{2}$National Institute of Healthcare Data Science, Nanjing University, China \\ 
    $^{3}$Alibaba Group, China \\
    {
        \tt \{yaozhao.nju,jchuang.nju\}@gmail.com, whu@nju.edu.cn,
    } \\ 
    {
        \tt \{qijin.cqj,xiaoxia.qxx,chengfu.huocf,afei\}@alibaba-inc.com
    }
 } 

\begin{document}
\maketitle
\begin{abstract}
Relation linking (RL) is a vital module in knowledge-based question answering (KBQA) systems. It aims to link the relations expressed in natural language (NL) to the corresponding ones in knowledge graph (KG). Existing methods mainly rely on the textual similarities between NL and KG to build relation links. Due to the ambiguity of NL and the incompleteness of KG, many relations in NL are implicitly expressed, and may not link to a single relation in KG, which challenges the current methods. 
In this paper, we propose an implicit RL method called ImRL, which links relation phrases in NL to relation paths in KG. To find proper relation paths, we propose a novel path ranking model that aligns not only textual information in the word embedding space but also structural information in the KG embedding space between relation phrases in NL and relation paths in KG. 
Besides, we leverage a gated mechanism with attention to inject prior knowledge from external paraphrase dictionaries to address the relation phrases with vague meaning. 
Our experiments on two benchmark and a newly-created datasets show that ImRL significantly outperforms several state-of-the-art methods, especially for implicit RL.
\end{abstract}

\section{Introduction}
\label{sect:intro}
The past few years have witnessed the rapid development of knowledge-based question answering (KBQA), which aims to answer natural language (NL) questions over knowledge graph (KG), e.g., DBpedia~\cite{DBpedia} and Freebase~\cite{Freebase}. In many KBQA systems~\cite{LinkingSurvey,NSQA}, as a fundamental module, relation linking (RL) seeks to link a relation phrase expressed in NL to a corresponding relation in KG, which has a great impact on the accuracy of subsequent steps like query construction and even the entire KBQA systems.

Previous studies~\cite{ReMatch,SIBKB,SimRel} focus on the similarities between the relation phrases and the text descriptions (e.g., local names) of relations in KG, and use the textual measures to link them. We refer to these studies as \emph{explicit} RL, because they all assume that the relations in NL are explicitly expressed and can be recognized by existing NLP tools.
However, according to~\cite{Falcon}, while existing RL methods perform well on several benchmark datasets, they are still challenged by \emph{implicit} relations prevalent in the real world.

\begin{figure}[t]
\centering
\includegraphics[width=1\columnwidth]{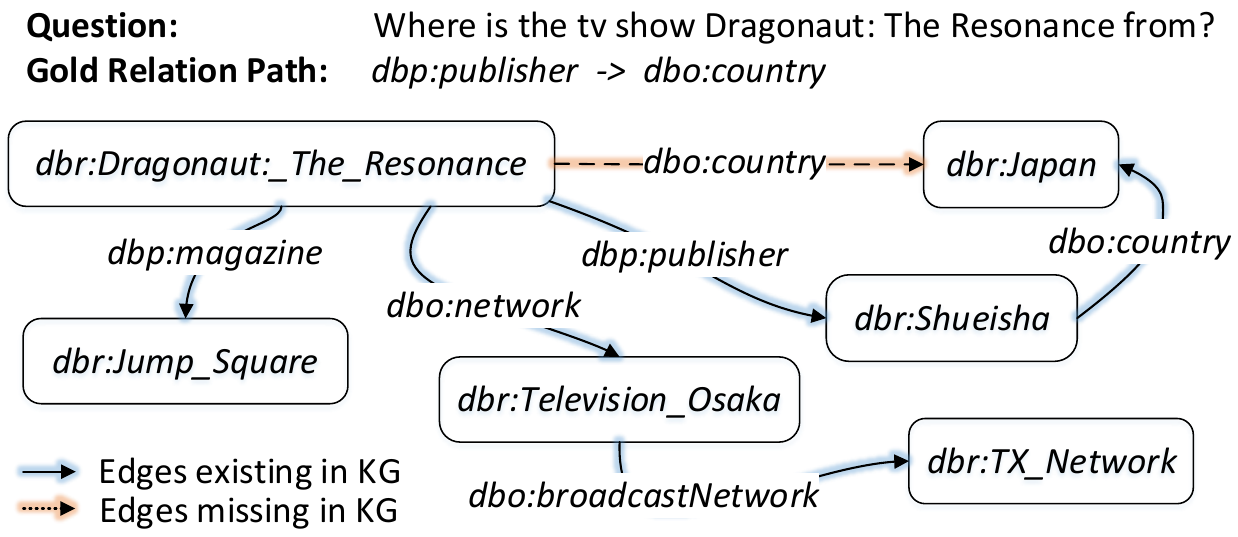}
\caption{An example of RL to DBpedia. There is no explicit relation between \textit{dbr:Dragonaut:\_The\_Resonance} and \textit{dbr:Japan}. We expect to implicitly link the phrase ``from'' to an indirect relation path \textit{dbp:publisher} $\rightarrow$ \textit{dbo:country}.
}
\label{fig1}
\end{figure}

Figure~\ref{fig1} illustrates a motivating example derived from \cite{SimpleQuestionsDBpedia}.
Assuming that the entity phrase ``Dragonaut: The Resonance'' has already been linked to \textit{dbr:Dragonaut:\_The\_Resonance},\footnote{We denote entities and relation phrases in NL by ``quotation marks'', and entities and relations in KG by \textit{italics}. \textit{dbr:}, \textit{dbo:} and \textit{dbp:} refer to \url{http://dbpedia.org/resource/}, \url{http://dbpedia.org/ontology/} and \url{http://dbpedia.org/property/}, respectively.} a typical method, e.g., 
EARL \cite{EARL}, conducts RL with three steps: detecting the relation phrases in the question, generating the candidate relations in KG according to each phrase, and ranking them by calculating the similarities. 
However, handling the above question faces two challenges: (1) KG (e.g., DBpedia) is incomplete, which leads to the situation that there is no direct relation between the entity \textit{dbr:Dragonaut:\_The\_Resonance} and the answer \textit{dbr:Japan}. Therefore, it is only possible to link them through an implicit relation path expressing a similar meaning as \textit{dbo:country}. (2) Due to the diversity of NL, there does not exist any explicit relation phrase except the word ``from'', but ``from'' has a vague meaning and cannot do much for disambiguation. 
Obviously, the expressions of relation phrase ``from'' and relation path \textit{dbp:publisher} $\rightarrow$ \textit{dbo:country} in this example are very different, which leads to the poor performance of existing methods, since they rely on the textual similarity. Besides, according to the statistics in~\cite{MPE4KBC}, 93.8\% of people in Freebase do not have \textit{place of birth}, and 78.5\% of people do not have \textit{nationality}, which shows that the situation of RL through implicit relation paths may occur in question answering frequently. 

In this paper, we focus on the problem of \emph{implicit RL}, where an implicit relation means the one not explicitly expressed in NL or cannot be linked to a single relation in KG. This is different from the studies, e.g., \cite{Review_add}, which use multiple explicit relations for multi-hop reasoning. To address such implicitness, we propose a new RL method called \textbf{ImRL} that links relation phrases in NL to relation paths in KG. 
To find proper relation paths, we propose a novel path ranking model that captures both textual information in the word embedding space and structural information in the KG embedding space, and align them between relation phrases in NL and relation paths in KG.
In this way, in addition to the textual information from word embeddings, ImRL can also use the structural information possessed by KG embeddings to capture the correlation between single relations and relation paths, thereby expressing the missing relations through similar relation paths in the incomplete KG.
Considering the work~\cite{Paraphrase} that constructs a dictionary of paraphrases and establishes a mapping relationship between NL phrases and relations in KG, which can be prior knowledge for implicit relation phrases with vague meanings, ImRL leverages a gated mechanism with attention to inject prior knowledge from external dictionaries into the model, and integrates all the information to predict the links between implicit relation phrases and relations (or relation paths) in KG.

In summary, our contributions are threefold:
\begin{compactitem}
\item We propose a novel ranking model, with the aim to eliminate the influence of incomplete KG. It aligns the textual information in the word embedding space and the structural information in the KG embedding space between relation phrases in NL and relation paths in KG. 
\item We explore a variety of ways to inject the knowledge in a paraphrase dictionary into the model, and choose a gate-based method with attention mechanism for knowledge injection, which can provide auxiliary information for the relation phrases with vague meanings.
\item We conduct experiments on three datasets, including two benchmark datasets and a newly-created dataset that particularly focuses on implicit RL. Our results show that ImRL outperforms several state-of-the-art competitors, especially for linking implicit relations. 
\end{compactitem}

\section{Related Work}
\label{sect:relat}
Existing RL approaches mainly focus on two tasks: relation candidate generation and relation disambiguation. 
For the first task, ReMatch~\cite{ReMatch}, SIBKB~\cite{SIBKB} and EERL~\cite{EERL} generate candidates by running a textual similarity-based method over pre-built dictionaries constructed by analyzing the NL patterns contained in massive text corpora through frequent item mining or crowdsourcing. Several widely-used dictionaries include PATTY~\cite{PATTY}, PPDB~\cite{PPDB}, Paraphrase~\cite{Paraphrase}, etc.

As for the second task, EARL~\cite{EARL} leverages the connection density of KG for ranking. Falcon~\cite{Falcon} uses several fundamental principles of English morphology to obtain auxiliary information, and exploits an alignment model for linking. With the development of deep learning, a few deep learning-based methods appear. SLING~\cite{SLING} leverages abstract meaning representation (AMR) to capture the semantic relationships in a question, and ranks the candidates with a Transformer-based model, which is trained in a distantly supervised manner. SimReL~\cite{SimRel} also uses AMR, and trains a BERT-based model for relation disambiguation.
In addition to the aforementioned studies using DBpedia, there are also some studies, e.g., KBPearl~\cite{KBPearl} and Falcon 2.0~\cite{Falcon2.0}, towards other KGs such as Wikidata~\cite{Wikidata}. 

Despite some literature~\cite{EERL,Falcon,SLING} mentions the issue of implicit relations, there is still much left for improvement. For instance, EERL discovers the implicit relations by using the domains and ranges of  properties in KG, but it cannot handle the issue of RL through relation paths. Falcon and SLING mainly focus on the implicit relations in phrases expressing the meaning about country, e.g., ``Spanish movie''. However, they perform poorly for other kinds of implicit relations, and cannot handle RL through relation paths either.

In recent years, KG embedding has become a popular area of research~\cite{KGESurvery}. Its main idea is to encode entities and relations in KG into a continuous low-dimensional embedding space. 
These embeddings can be further applied to various tasks, such as KG completion and KBQA. Many studies in the KG completion field \cite{TransE,PRA,PR_KGC_1} predict the possibility of the establishment of a specific triple through the embedding of entities and relations. 
Recently, researchers in the KBQA field have also begun to leverage KG embeddings. The studies~\cite{KGEQA,KBEQA} try to learn a deep model that maps a NL question into a KG embedding space, and combine the KG embeddings of topic entities in the question with the question embedding to infer the answer. 
To the best of our knowledge, there is no prior work focusing on using KG embedding for implicit RL.

\begin{figure*}[t]
\centering
\includegraphics[width=1\textwidth]{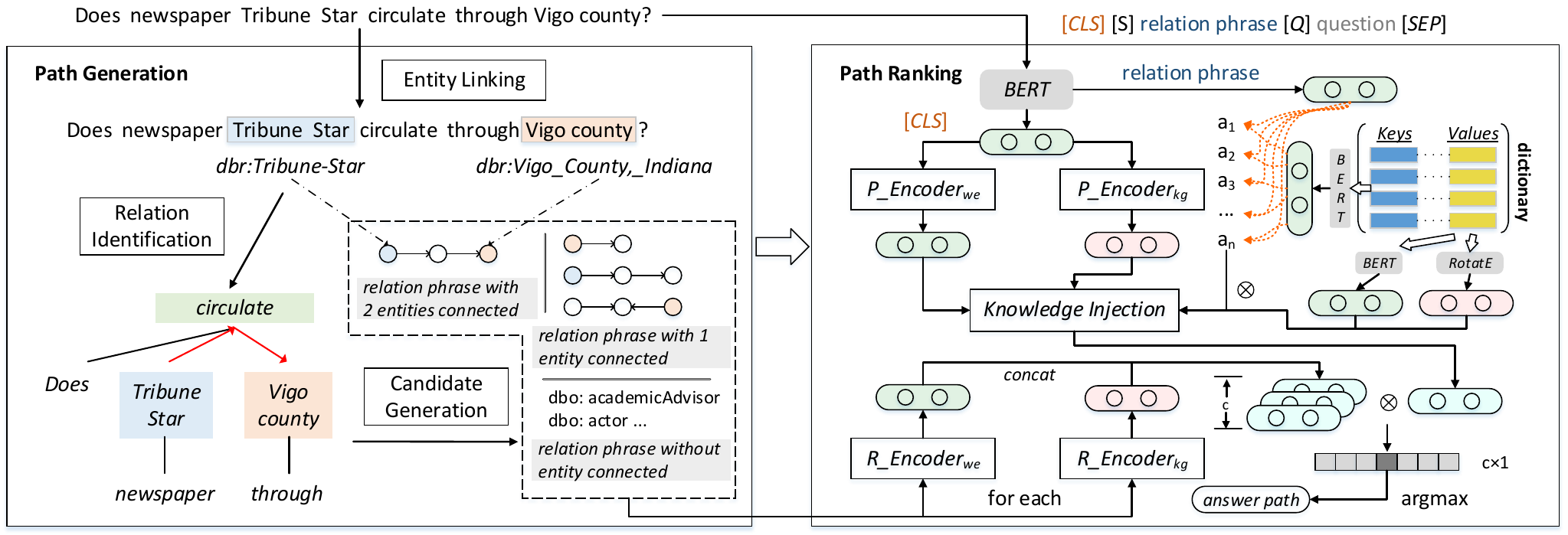} 
\caption{Overview of our method.
The method has two parts: (1) \emph{Path generation} parses the input question and finds the relation path candidates in the KG, by entity linking, relation identification and candidate generation. (2) \emph{Path ranking} encodes the relation phrase in the question and path candidates in the KG in the BERT embedding space and RotatE embedding space, utilizes a ranking model to rank those candidates, and takes the one with the highest similarity score as answer.
It also leverages a gated mechanism with attention to inject prior knowledge from external dictionaries to help relation disambiguation.
}
\label{fig2}
\end{figure*}

\section{Implicit Relation Linking}
\label{sect:method}

A knowledge graph (KG), denoted by $\mathcal{G}$, can be defined as a triple $\mathcal{G}=(\mathcal{E},\mathcal{R},\mathcal{T})$, where $\mathcal{E}$ is the entity set, $\mathcal{R}$ is the relation set, and $\mathcal{T}\subseteq\mathcal{E}\times\mathcal{R}\times\mathcal{E}$ denotes the relational triple set. Each entity in $\mathcal{E}$ is denoted by $e$, and each relation in $\mathcal{R}$ is denoted by $r$. If there exists a set of $m$ triples $\{(e_0,r_1,e_1), \dots, (e_{m-1},$ $r_m,e_m)\} \subseteq \mathcal{T} $, we say that there is an $m$-hop relation path $p$ between $e_0$ and $e_m$, denoted by $p=(r_{1},\dots,r_{m})$. The goal of RL can be formulated as: Given a NL question $q$ and a KG $\mathcal{G}$, linking a relation phrase $s$ in $q$ to its corresponding relation $r$ or relation path $p$ in $\mathcal{G}$. 
Figure~\ref{fig2} shows the architecture of our method.

\subsection{Path Generation}
\label{sect:path-generation}
The path generation module consists of three steps: 

\smallskip
\noindent\textbf{Entity linking}
aims to link the phrases in NL that represent real-world objects to the corresponding entities in KG. 
Recent studies~\cite{LinkingSurvey} have shown that entity linking can affect RL. 
To reduce the influence of entity linking on RL, we use the existing tool Falcon \cite{Falcon} to link entities in advance, which has shown accurate results on entity linking (around 0.83 of F1-score on the LC-QuAD dataset as reported). The results are then fed into subsequent steps. 

\smallskip
\noindent\textbf{Relation identification}
\label{relation-identification}
is to determine whether there is any relation phrase in a question with entity connected to it.
Inspired by \cite{gAnswer}, which demonstrates that using rules to transform the dependency tree into a directed graph and using a heuristic search to find the shortest path between two nodes can effectively decompose the complex question into multiple simple triples, thereby discovering the relations. 
We use a similar method to identify the relations in a question.
Furthermore, we use a dictionary-based method~\cite{CL} to identify those phrases that can be linked to classes in KG. In particular, wh-words, e.g., ``what'' and ``when'', are also regarded as class phrases.
For each pair of nodes in the graph, we consider there is a relation between them if and only if 
(1) both nodes are phrases linked to entities or classes;
(2) all intermediate nodes in the path between the nodes are not linked to entities or classes. 
The word sequence composed of intermediate nodes is obtained by the breadth-first-search (BFS) algorithm, and regarded as a relation phrases. For more details, please refer to Algorithm~\ref{algo:ri}.
In this way, we can obtain some relation phrases with entities connected to them. 
Furthermore, according to \cite{SLING}, AMR can effectively capture the semantic information at the sentence level, thus we use a similar method for relation phrase extension.
For example, for a phrase like ``German actor'', ``country'' can be recognized as the relation phrase contained in it through AMR. We refer the reader to \cite{SLING} for more details.

\label{sec:ri-alg}
\renewcommand{\algorithmicrequire}{\textbf{Input:}}  
\renewcommand{\algorithmicensure}{\textbf{Output:}}  
\begin{algorithm}[!h]
  \caption{Relation Identification}  
  \begin{algorithmic}[0]
    \Require
    $Q$: Natural language question;
    $EC$: Result set of entity and class linking;
    $subj\_relations$: [subj, nsubj, nsubjpass, csubj, csubjpass, xsubj, poss, partmod];
    $obj\_relations$: [obj, pobj, dobj, iobj];
    \Ensure
    Related entities/classes and relation between them;
    \State Use dependency analysis to obtain an undirected tree $T$;
      \For{$ec_i$ in $EC$}
      \State Combine nodes representing $ec_i$ in $T$ into one;
      \State Delete all the edges which are related to the combined node, update $T$;
      \EndFor
    \State Build a new empty graph $G$, copy nodes in $T$ to $G$;
    \For{$t_i$ in $T$}\Comment{$t_i$ stands for $(node_1,r,node_2)$}
      \If {$r$ in $subj\_relations$}
      \State Add an edge from node1 to $node_2$, weight is the number of entity/class between $node_1$ and $node_2$;
      \ElsIf {$r$ in $obj\_relations$}
      \State Add an edge from $node_2$ to $node_1$;
      \Else
      \State Add both edges mentioned above;
      \EndIf
      \EndFor
      \State Update distance between nodes with the Floyd shortest path algorithm;
      \For{$node_i,node_j$ in $G$}
      \If{$distance(node_i,node_j)=2$ and $node_i,$ $node_j$ are both entities/classes}
      \State Use the BFS algorithm to obtain the relation $r$ between them, return $(node_i,r,node_j)$;
      \EndIf
      \EndFor
  \end{algorithmic}
  \label{algo:ri}
\end{algorithm}

\smallskip
\noindent\textbf{Candidate generation.}
Existing RL methods only consider single relations when generating candidates. However, KG is widely acknowledged as incomplete \cite{KGESurvery}, which makes it fail to find the corresponding single relations in KG for some phrases and needs to be expressed by relation paths. We extend the RL task to link phrases to relation paths, where conventional RL can be regarded as a simpler case when the path is one hop.
We enumerate the relation path candidates in different ways according to the number of connected entities of a phrase: (1) If a relation phrase is connected with two entities, all the paths between them are regarded as candidates. (2) If there is only one entity, all the paths within \textit{$M$}-hop are considered as candidates. (3) Otherwise, if there is any phrase without any entity association, all possible common single relations in KG are considered as candidates.
In order to effectively reduce the input size of subsequent modules, we select $c$ candidate paths whose local names are most similar to the relation phrases, where $c$ is a hyperparameter and the similarity is calculated with cosine similarity between the word embeddings of phrases and local names.

\subsection{Path Ranking}
\label{path-ranking}
We propose a novel ranking model to find the relation path most relevant to the given relation phrase. Different from those studies using path ranking algorithm for KG completion \cite{PR_KGC_1, PR_KGC_2, PR_KGC_3, MINERVA}, which leverage the entities and relations in the KG for path reasoning, our path ranking model additionally considers the relation phrase in the question for ranking.

\smallskip
\noindent\textbf{Relation path encoder.}
The relations in KG can provide rich information. Most existing RL methods mainly focus on the local names of KG relations, and leverage language models to capture the semantics. However, they ignore the structural information in massive triples. In order to make use of both information, we use this relation path encoder to model relation paths in the word embedding space, denoted by $R\_Encoder_{
we}$, and the KG embedding space, denoted by $R\_Encoder_{kg}$.

$R\_Encoder_{we}$ consists of a BERT encoder \cite{BERT} and a feed-forward neural network (FFNN). For each relation path candidate $p=(r_{1},\dots,r_{m})$, the sequence of local names $input_{p}$ of $p$ is denoted by $\langle[CLS],name_{r_1},$ $[SEP],\dots, name_{r_{m}}, [SEP]\rangle$, which is fed to the encoder to obtain the embedding $\mathbf{p}_{we} \in \mathbb{R}^{d_{we}}$:
\begin{align}
    \mathbf{p}_{we} = \sigma(\mathbf{W}_{1} \textsc{Bert}(input_{p}) + \mathbf{b}_{1}),
\end{align}
where $\sigma()$ is the ReLU activation, $\mathbf{W}_{1}$ is the weight matrix of FFNN, and $\mathbf{b}_{1}$ is the bias vector.

$R\_Encoder_{kg}$ uses a KG embedding model to encode the relations in KG. Considering that RotatE~\cite{RotatE} has been proven to achieve good results in existing work~\cite{KGEQA}, and have the properties of composition and symmetry of relations, we choose it as the KG embedding model. According to \cite{KGESurvery}, KG embeddings have a certain ability for reasoning. Thus, the KG embedding of a relation path with $d_{kg}$ dimension $\mathbf{p}_{kg} \in \mathbb{R}^{d_{kg}}$ can be obtained by merging all the KG embeddings of single relations within the path through composition operation:
\begin{align}
    \mathbf{p}_{kg} &= \textsc{RotatE}(r_1) \circ \cdots \circ \textsc{RotatE}(r_m),
\end{align}
where $\circ$ denotes the Hadamard product.

\smallskip
\noindent\textbf{Phrase encoder.}
In order to compare with the relation paths in KG equally, the relation phrases in NL questions also need to be encoded into two spaces to capture different aspects of information.

$P\_Encoder_{we}$ aims to encode the relation phrase $s$ with its context in a question $q$ into an embedding $\mathbf{s}_{we} \in \mathbb{R}^{d_{we}}$. It has the same architecture as $R\_Encoder_{we}$. The difference is that the input sequence is added with two special tokens $[S]$ and $[Q]$ to separate the relation phrase and its question context, i.e., $input_{s} = \langle[CLS],[S],s,[Q],q,[SEP]\rangle$:
\begin{align}
    \mathbf{s}_{we} &= \sigma(\mathbf{W}_{2} \textsc{Bert}(input_{s}) + \mathbf{b}_{2}),
\end{align}
where $\mathbf{W}_{2}$ and $\mathbf{b}_{2}$ are two learnable parameters.

$P\_Encoder_{kg}$ aims to encode the relation phrase and the question into a KG embedding $\mathbf{s}_{kg} \in \mathbb{R}^{d_{kg}}$. It first uses BERT to encode the whole $input_{s}$, and pick the embedding of $[CLS]$, which is then fed to a multi-layer perceptron $\textsc{mlp}_{we\rightarrow kg}$ with a 3-layer FFNN and ReLU activation to obtain $\mathbf{s}_{kg}$.
In a question, a relation phrase $s$ creates a connection between entities, and we expect the KG embedding $\mathbf{s}_{kg}$ to reflect the KG embedding of the relation path in KG connecting these entities, i.e., $\mathcal{\phi}(\mathbf{e}_{1}, \mathbf{s}_{kg}, \mathbf{e}_{2})\approx 0$, where $\mathcal{\phi}$ denotes the score function of RotatE, and $\mathbf{e}_{1},\mathbf{e}_{2}$ are the KG embeddings of two connected entities. This is equivalent to directly using the KG embedding of relation path between the two entities to train the model.
Thus, during the training phase, we use the KG embedding of relation path $\mathbf{p}_{kg}$ corresponding to the relation phrase $s$ to guide the learning of $\mathbf{s}_{kg}$, and the loss function is mean-square error loss: 
\begin{align}
    \mathcal{L}_{\alpha} &= \frac{1}{|\mathcal{D}|} \sum_{(s,q,p) \in \mathcal{D}}^{} \left\|\mathbf{s}_{kg} - \mathbf{p}_{kg}\right\|_2^2,
\end{align}
where $D$ is the training set, $\mathbf{s}_{kg}$ is the KG embedding of relation phrase $s$, $\mathbf{p}_{kg}$ is the KG embedding of relation path $p$, and $\|\cdot\|_2$ is the $L_2$-norm.

\smallskip
\noindent\textbf{Knowledge injection.}
The previous works \cite{PATTY,Paraphrase} pre-collect some paraphrase pairs (e.g., ``be born in'' is a paraphrase for relation \textit{hometown}) between relation phrases in NL and KG relations through frequent item mining or crowdsourcing. These paraphrases can be considered as prior knowledge about the diverse meanings of NL. For example, if we have some prior examples of linking ``in'' to KG relations, e.g., \textit{locatedIn}, we can use such knowledge to enhance the meanings of implicit relation phrases ``in''. In this paper, we choose a predicate paraphrase dictionary called Paraphrase \cite{Paraphrase} as the knowledge source. Paraphrase maps NL patterns to DBpedia relations, which currently covers 2,284 relations and 31,130 paraphrase pairs.

We regard the NL patterns in Paraphrase as \textit{Keys}, and corresponding relations as \textit{Values}. \textit{Keys} provide textual knowledge from the perspective of NL, while \textit{Values} provide both textual and structural knowledge of NL and KG.
Same as the previous step, \textit{Keys} is encoded as $\mathbf{K} \in \mathbb{R}^{d_e \times d_{we}}$ using BERT, \textit{Values} is encoded as $\mathbf{V}_{we} \in \mathbb{R}^{d_e \times d_{we}}$ and $\mathbf{V}_{kg} \in \mathbb{R}^{d_e \times d_{kg}}$ using BERT and RotatE, respectively. Here, $d_e$ denotes the number of most similar paraphrase pairs to be selected. In this way, given a relation phrase, which is then encoded by BERT as $\mathbf{Q} \in \mathbb{R}^{1 \times d_{we}}$, we can use the attention mechanism \cite{Attention} to find the most relevant NL patterns in Paraphrase and integrate their corresponding knowledge into the original embedding:
\begin{align}
    \mathbf{k}_{we} &= \textsc{softmax}(\frac{\mathbf{Q}\mathbf{K}^T}{\sqrt{d_{we}}})\mathbf{V}_{we}, \\
    \mathbf{k}_{kg} &= \textsc{softmax}(\frac{\mathbf{Q}\mathbf{K}^T}{\sqrt{d_{kg}}})\mathbf{V}_{kg}.
\end{align}

We also explore a variety of ways to inject the knowledge into our model. Taking the fusion of $\mathbf{p}_{kg}$ and $\mathbf{k}_{kg}$ as an example, we use the following three methods:
\begin{compactitem}
    \item \textbf{Element-wise mean (MEAN)} is the most straightforward way to add two embeddings and take the average over each element:
    \begin{align}
        \mathbf{y}_{kg} &= \frac{1}{2}(\mathbf{s}_{kg} + \mathbf{k}_{kg}) .
    \end{align}

    \item \textbf{Concatenation (CAT).} By concatenating the two embeddings and feeding them to a linear layer, all the information can be reserved:
    \begin{align}
        \mathbf{y}_{kg} &= \mathbf{W}_3[\mathbf{s}_{kg} \,;\, \mathbf{k}_{kg}]+\mathbf{b}_3,
    \end{align}
    where $;$ denotes the concatenation operation. $\mathbf{W}_3$ and $\mathbf{b}_3$ are two learnable parameters.
    
    \item \textbf{Gated mechanism (GATE).} To retain the valuable original information and absorb new knowledge, a gated mechanism can be used for a trade-off between them:
     \begin{align}
        g &= \delta(\mathbf{W}_4[\mathbf{s}_{kg} \,;\, \mathbf{k}_{kg}]+\mathbf{b}_4), \\
        \mathbf{y}_{kg} &= g \cdot \mathbf{s}_{kg} + (1-g) \cdot \mathbf{k}_{kg} ,
    \end{align}
where $\delta()$ denotes the sigmoid activation. $\mathbf{W}_4$ and $\mathbf{b}_4$ are two learnable parameters.
\end{compactitem}

Then, given two fused embeddings, $\mathbf{y}_{we}$ and $\mathbf{y}_{kg}$ of a relation phrase $s$ in question $q$, and the embeddings of a relation path candidate $p$, $\mathbf{r}_{we}$ and $\mathbf{r}_{kg}$, the score function $\psi()$ can be formulated as
    \begin{align}
        \psi(s,q,p) &= [\mathbf{y}_{we}\,;\,\mathbf{y}_{kg}] \cdot [\mathbf{p}_{we}\,;\,\mathbf{p}_{kg}].
    \end{align}

We select the cross entropy loss between the one-hot vector of ground truth and the score vector:
    \begin{align}
        \mathcal{L}_{\beta} = -\sum_{(s,q,p) \in \mathcal{D}}\log \frac{e^{\psi(s,q,p)}}{ e^{\psi(s,q,p)} + \sum\limits_{p'\in \mathcal{N}_p} e^{\psi(s,q,p')}},
    \end{align}
where $\mathcal{N}_{p}$ denotes the set of negative relation paths obtained by negative sampling based on $p$. The negative sampling is implemented by randomly sampling $N$ relations or relation paths in KG which are different from $p$.

Finally, the overall loss function is defined as
    \begin{align}
        \mathcal{L} = \mathcal{L}_{\alpha} + \lambda \mathcal{L}_{\beta}.
    \end{align}

\section{Experiments and Results}
\label{sect:exper}

\subsection{Dataset Construction}
Due to the high correlation between RL and KBQA, we follow the works \cite{SLING,SimRel} to reuse the KBQA datasets as the benchmarks to evaluate the performance of RL.
We briefly introduce them as follows:

\begin{compactitem}
\item \textbf{LC-QuAD 1.0}~\cite{LC-QuAD} is a large complex questions dataset for KBQA, which contains 5,000 questions and corresponding SPARQL queries over DBpedia (version 2016-04). Only 1\% of the questions involve implicit relation links.
\item \textbf{QALD-9}~\cite{QALD-9} is a popular benchmark dataset for KBQA over DBpedia. It provides 408 training questions and 150 test questions, 6\% of which need implicit RL.
\item \textbf{Path-based SimpleQuestions dataset} (PathSQ) is a new dataset constructed by ourselves for implicit RL evaluation. It is based on DBpedia (version 2016-04) and contains 400 questions that need to be linked through a two-hop relation path. Table~\ref{tab:PathSQ-example} lists an example, and we explain its construction details below.
\end{compactitem}

\smallskip
\noindent\textbf{Construction of PathSQ.} 
Existing benchmarks contain few implicit relations and cannot be used to evaluate implicit RL well.
As for other KBQA datasets, e.g., ComplexWebQuestions \cite{ComplexWebQuestions}, the proportion of implicit relations is also small.
Thus, we decide to construct a new dataset by ourselves for implicit RL evaluation.

In SimpleQuestions \cite{SimpleQuestions}, some questions have corresponding single relations in Freebase but need to be expressed using multiple relations in DBpedia. Based on this observation, we collect the questions in SimpleQuestions that must map each relation phrase in NL to a relation path in DBpedia. 
We only consider two-hop relation paths, since paths with more relations are generally rare and spurious \cite{SimpleQuestionsDBpedia}. We construct PathSQ with the following three steps: (1) We remove the question and answer pair in SimpleQuestions such that the entity in the question or the answer entity cannot be mapped to DBpedia. We use the officially released file\footnote{\scriptsize\url{http://downloads.dbpedia.org/2016-10/core-i18n/en/freebase_links_en.ttl.bz2}} to achieve this goal, which contains the entity mappings between Freebase and DBpedia. (2) For each remaining question and answer pair, we construct a SPARQL query to determine whether there exists a direct relation between the entity in the question and the answer entity. If exists, we remove this pair. (3) We construct another SPARQL query to obtain all the two-hop paths between the two entities for each remaining pair. Among all the paths, we manually label the correct ones and try our best effort to mine 400 questions for experiments.

\begin{table}[t]
    \centering
    \small\begin{tabular}{l|p{5cm}}
    		\toprule 
    		    Question & What is Claire Stansfield's nationality?\\
    		\midrule
    		    Head entity & \textit{dbr:Claire\_Stansfield}\\
    		    Relation phrase & nationality\\
    		    Implicit relation & \textit{dbo:birthPlace} $\rightarrow$ \textit{dbo:country}\\
    		    Answer & \textit{dbr:United\_Kingdom}\\
    		\bottomrule 
	\end{tabular}
    \caption{An example in the PathSQ dataset.}
    \label{tab:PathSQ-example}
\end{table}

\begin{table*}[t]
\centering
\small\begin{tabular}{p{0.3cm} | p{1.6cm}<{} | p{1cm}<{\centering} p{1cm}<{\centering} p{1cm}<{\centering} p{1cm}<{\centering} p{1cm}<{\centering} p{1cm}<{\centering} p{1cm}<{\centering} p{1cm}<{\centering} p{1cm}<{\centering}}
	\toprule
	\multicolumn{2}{l|}{\raisebox{-0.8\normalbaselineskip}[0pt][0pt]{Methods}} & \multicolumn{3}{c}{LC-QuAD} & \multicolumn{3}{c}{QALD-9} & \multicolumn{3}{c}{PathSQ} \\
	\cmidrule{3-11} 
	\multicolumn{2}{c|}{} & P$\,\uparrow$ & R$\,\uparrow$ & F1$\,\uparrow$ & P$\,\uparrow$ & R$\,\uparrow$ & F1$\,\uparrow$ & P$\,\uparrow$ & R$\,\uparrow$ & F1$\,\uparrow$ \\
	\midrule
    \raisebox{-2\normalbaselineskip}[0pt][0pt]{\rotatebox[origin=c]{90}{\ \ \ \ Textual}}
        & SIBKB   & 0.13 & 0.15 & 0.14 & 0.14 & 0.10 & 0.11 & 0.23 & 0.12 & 0.16 \\
        & ReMatch & 0.15 & 0.17 & 0.16 & 0.14 & 0.15 & 0.14 & 0.24 & 0.15 & 0.18 \\
        & EARL    & 0.17 & 0.21 & 0.18 & 0.15 & 0.17 & 0.16 & 0.06 & 0.05 & 0.05 \\
        & Falcon  & 0.42 & 0.44 & 0.43 & 0.23 & 0.23 & 0.23 & 0.30 & \underline{0.19} & 0.23 \\
    \midrule    
    \raisebox{-1.75\normalbaselineskip}[0pt][0pt]{\rotatebox[origin=c]{90}{\ \ \ \ Neural}}
        &  SLING & 0.41 & \textbf{0.55} & 0.47 & 0.39 & \textbf{0.50} & 0.44 & 0.09 & 0.05 & 0.07       \\
        & SimReL & \underline{0.51} & \underline{0.51} & \underline{0.51} & \underline{0.46} & 0.44 & \underline{0.45} & \underline{0.39} & \underline{0.19} & \underline{0.26}       \\
    \cmidrule{2-11}
        & ImRL (ours)  & \textbf{0.59} & 0.49 & \textbf{0.54} & \textbf{0.51} & \underline{0.46} & \textbf{0.48} & \textbf{0.46} & \textbf{0.41} & \textbf{0.43}       \\
	\bottomrule
\end{tabular}
\caption{Relation linking performance comparison. We mark the best scores in \textbf{bold} and second-best \underline{underlined}.}
\label{tab:result}
\end{table*}

\subsection{Comparative Methods}
We compare our ImRL with six state-of-the-art methods. All of them support RL over DBpedia.
\begin{compactitem}
\item \textbf{SIBKB} \cite{SIBKB} uses PATTY as the underlying knowledge source, and establishes a bipartite graph index from common phrases in NL to KG relations for RL.
\item \textbf{Rematch} \cite{ReMatch} exploits WordNet and dependency parsing to model relation phrases with their underlying parts of speech, and uses various similarity metrics for ranking.
\item \textbf{EARL} \cite{EARL} models the linking problem as the generalized traveling salesman problem, and uses connection density to link entities and relations jointly.
\item \textbf{Falcon} \cite{Falcon} enhances entity and relation linking through cross-KG entity and relation alignment and basic principles of English morphology.
\item \textbf{SLING} \cite{SLING} leverages AMR for preprocessing, trains a Transformer-based model based on distant supervision, and integrates several complementary signals for RL.
\item \textbf{SimReL} \cite{SimRel} also uses AMR for question representation, and employs a Siamese network with BERT to link relations in an end-to-end manner.
\end{compactitem}

\subsection{Experiment Setup}
\label{exper-setup}
We implement ImRL in Python 3. All experiments are conducted on a workstation with two Xeon Silver 4215R CPUs, 384GB memory and an NVIDIA TITAN RTX graphics card. We use the API of Falcon \cite{Falcon} to link entities to DBpedia, and use StoG \cite{StoG} to generate AMR graphs in relation identification.
For word embeddings, we use the pre-trained BERT-base model~\cite{HuggingFace} with $d_{we}=768$ to initialize the parameters and freeze them in the experiments. 
For KG embeddings, we extract 200,000 popular entities and their related triples from DBpedia as the training corpus, and employ RotatE with $d_{kg}=200$ for embedding learning. 

As for the hyperparameters, the maximum number of relation path candidates is empirically set to $c=30$, the maximum length of path candidates is set to $M=2$, the number of negative sampling relations is set to $N=29$, and the number of selected paraphrase pairs is set to $d_e=10$. For all the experiments, we assign the learning rate to $3e^{-4}$ and the batch size to 256. The weight $\lambda$ in loss function is set to 1.
The previous works evaluate the performance of RL with precision, recall and F1-score. For consistency, we use the same metrics in our experiments. For LC-QuAD and QALD-9, we use the training set to train the model, and evaluate on the testing set. For PathSQ, due to the small scale, we treat all the dataset as the testing set, and use the training sets of LC-QuAD and QALD-9.

\subsection{Main Results}
\label{main-results}
Table~\ref{tab:result} lists the results of all the methods.
From the table, ImRL achieves the state-of-the-art results on LC-QuAD and QALD-9.

Take QALD-9 for example, 94\% of the questions are with explicit single relations. For those questions, the F1-score of ImRL can reach 0.48, showing that even without a specific design for explicit RL, ImRL still performs well for it. The reason is that leveraging word embeddings and KG embeddings is common to both implicit and explicit relations. For explicit relations, since they usually contain a single relation, both $R\_{Encoder}$ and $P\_{Encoder}$ only need to encode one relation phrase or relation local name, which makes the matching of word embeddings easier. In addition, due to the use of KG embeddings and external knowledge of paraphrase dictionary as additional information, ImRL outperforms the latest method SimReL without the two information. However, since the path generation may cause correct relation paths missing in the candidate set, the recall of ImRL is lower than precision.

For the remaining 6\% questions with implicit relations in QALD-9, the F1-score of our method can reach 0.47. In order to further verify the performance of ImRL on implicit relations, we conduct another experiment on PathSQ. As shown in Table~\ref{tab:result}, ImRL dominates in terms of all the three evaluation metrics, while the other methods obtain rather poor results. This is because all the questions in PathSQ need to be linked through implicit relation paths, which is challenging for the existing methods. By contrast, ImRL can achieve the reasoning of relation paths according to the composition property of KG embeddings, thereby effectively improving the accuracy of implicit RL. 

Furthermore, ImRL can also handle unseen relations in the inference phase. Taking QALD-9 as an example, there are 46 questions in the testing set whose relations have never appeared in the training set. Still, ImRL can correctly predict 10 of them, which shows the generalization of ImRL.

\subsection{Detailed Analysis}
\label{sect:detailed_analysis}
We conduct more experiments to validate the proposed method. All the following experiments are based on QALD-9 dataset. 

\begin{table}[!t]
 \centering
 \small\begin{tabular}{l|ccc}
  \toprule Methods & P$\,\uparrow$ & R$\,\uparrow$ & F1$\,\uparrow$ \\
	\midrule 
	ImRL & \textbf{0.51} & \textbf{0.46} & \textbf{0.48} \\
	w/o KG embeddings & 0.46 & 0.42 & 0.44 \\
	w/o word embeddings & 0.42 & 0.37 & 0.39 \\
	w/o KG and word embeddings$^*$ & 0.28 & 0.26 & 0.27 \\
  \bottomrule
  \multicolumn{4}{l}{$*$ denotes using string matching without embeddings.}
 \end{tabular}
 \caption{Results of different embedding models.}
 \label{tab:exp-embedding-model}
\end{table}
 
\smallskip
\noindent\textbf{Ablation study of embedding models.}
To measure the contribution of each embedding model in ImRL, we remove the module of obtaining KG embeddings and word embeddings in turn for ablation study. 
By removing the use of KG embeddings, the F1-score drops by 0.04, while by removing the use of word embeddings, it drops by 0.09. This shows that in our model, the word embeddings play a more important role than the KG embeddings. We believe that the root cause to the phenomenon is that pre-trained language models, e.g., BERT, are trained on huge corpora with several downstream tasks, thus the word embeddings can provide more prior knowledge. While the KG embedding model used by ImRL, i.e., RotatE, is trained on a small-scale corpus, and the obtained embeddings more focus on limited structural information, which leads to that the KG embeddings provide relatively less information than the word embeddings.

\smallskip
\noindent\textbf{Different knowledge injection methods.}
We explore three methods of injecting external knowledge from the paraphrase dictionary into ImRL, and compare them with the method without using the dictionary. The result is shown in Table~\ref{tab:exp-knowlege-injection}. All methods with injection achieve better results than those without it, which verifies the effectiveness of knowledge injection. Among all the methods, ImRL with a gated mechanism achieves the best results. This shows that the knowledge in the dictionary may also bring noises, thus by combining part of the external knowledge and the original embeddings of the phrases, more accurate external knowledge can be supplemented to enhance the representations of implicit relation phrases. It should be mentioned that utilizing paraphrase dictionaries also has certain limitations. For instance, different dictionaries target different KGs, and may need to be adjusted according to the schema of the KG.

\begin{table}[!t]
 \centering
 \small\begin{tabular}{l|ccc}
  \toprule Methods & P$\,\uparrow$ & R$\,\uparrow$ & F1$\,\uparrow$ \\
  \midrule 
    ImRL (GATE) & \textbf{0.51} & \textbf{0.46} & \textbf{0.48} \\
    ImRL (MEAN) & 0.50 & 0.44 & 0.47 \\
    ImRL (CAT) & 0.48 & 0.43 & 0.46 \\
    ImRL (None) & 0.47 & 0.42 & 0.45 \\
  \bottomrule
 \end{tabular}
 \caption{Results of different injection methods.}
 \label{tab:exp-knowlege-injection}
\end{table}

\smallskip
\noindent\textbf{Error propagation in pipeline.}
In order to reduce the impact of previous steps, and only evaluate the performance of the ranking model, we conduct an experiment assuming that the input to the Path Ranking module is the gold standard. That is, the correct relation path is always in relation path candidate set, which is the input of the Path Ranking module. Under this assumption, the F1-score of the model can reach 0.57, which shows that the aforementioned steps have a great impact on the model. Meanwhile, this also shows that the gap between the relations expressed in NL and those in KG is quite large. Even if lots of additional information is considered, it is still difficult to pick the most correct one from a set of relation path candidates, which reveals the challenge of RL task.

\smallskip
\noindent\textbf{Improvement to KBQA performance.}
As a vital module of KBQA, it is necessary to integrate ImRL into some existing KBQA systems for validation. We conduct the experiment on the most popular KBQA dataset LC-QuAD~\cite{LC-QuAD}. We choose gAnswer~\cite{gAnswer} as the baseline KBQA system, and replace its RL module with ImRL, which results in about 3.2\% increase of F1-score, verifying that our method can indeed help the existing KBQA systems. 
As a reference, SimReL reports its F1-score improvement on LC-QuAD as 2.4\% \cite{SimRel}.
It is worth noting that the same process can also be migrated to other KGs, e.g., Freebase, for further use.

\smallskip
\noindent\textbf{Case study.}
Table~\ref{tab:case-study} shows two real examples to help understanding.
In the first case from PathSQ, the relation phrase ``involved in'' needs to be expressed using the relation path \textit{dbo:commander} $\rightarrow$ \textit{dbo:militaryBranch} according to the gold standard. 
The state-of-the-art method SimReL returns \textit{dbo:militaryBranch}, as it can only find the single relation whose meaning is most similar to the meaning of the whole question as the answer. Differently, ImRL leverages the path ranking model to find the path that best expresses the meaning of ``army ... involved in''.
In the second case, there is no entity in this question of QALD-9, which causes that SimReL cannot enumerate any candidate relations since it depends on the result of entity linking. Although the method of generating candidates of ImRL is more robust, the absence of entities also has a great impact on it, making it give the wrong answer \textit{dbo:origin}. Linking this relation requires understanding the meaning of the noun phrase, which is incapable for all existing RL methods including ImRL. We leave this in future work.

\begin{table}[!t]
    \centering
    \small\begin{tabular}{l|l}
		\toprule
		Case 1 & What army was involved in Siege of Clonmel?  \\ 
		\midrule
		 Gold & \textit{dbo:commander} $\rightarrow$ \textit{dbo:militaryBranch} \\
		 SimReL & \textit{dbo:militaryBranch} \\
		 ImRL & \textit{dbo:commander} $\rightarrow$ \textit{dbo:militaryBranch} \\
		\bottomrule
		Case 2 & Give me all animals that are extinct.  \\ 
		\midrule
		 Gold & \textit{dbo:conservationStatus}  \\
		 SimReL & \textit{null} \\
		 ImRL & \textit{dbo:origin} \\
		\bottomrule
	\end{tabular}
    \caption{Case study of linking results.}
    \label{tab:case-study}
\end{table}

\section{Conclusion}
\label{sect:conclusion}
In this paper, we propose ImRL, an implicit RL method that can better deal with the linking of implicit relations in NL and incomplete KG. We extend the RL task to implicit relation path linking, and propose a novel ranking model with knowledge injection. We evaluate our model on three datasets. The results show that our method achieves superior performance on LC-QuAD and QALD-9, where the relations are mostly explicit, and PathSQ, where the relations are all implicit, demonstrating that ImRL can not only deal with relation phrases that are not explicitly expressed in NL, but also perform relation path reasoning in KG. 
In future work, we will explore how to combine entity linking and implicit RL for improvement.

\section*{Acknowledgements}
This work is supported by National Natural Science Foundation of China (No. 61872172), and Alibaba Group through Alibaba Research Fellowship Program.

\bibliography{custom}
\bibliographystyle{acl_natbib}

\clearpage
\appendix

\section{Sensitivity Analysis of Candidate Path Number}
\label{sec:sa-c}

In order to assess the sensitivity of ImRL to the number of candidate paths, we report the empirical analyses using different numbers of candidate paths $c$ on QALD-9. The metrics are mean reciprocal rank (MRR), Hits@$c$ and F1-score. MRR represents the multiplicative inverse of the rank of the gold path in the candidate path list. Hits@$c$ indicates the proportion of the questions where the gold path ranks in the top-$c$ in the candidate list. F1-score represents the performance of ImRL on QALD-9.
As shown in Table~\ref{tab:sa-c}, with the increase of $c$, MRR, Hits@$c$ and F1-score all increase. However, when $c$ reaches 30, the increment of $c$ has little influence on the model performance. Thus, we choose to set $c=30$.

\begin{table}[!h]
 \centering
 \small\begin{tabular}{c|ccc}
  \toprule Candidate path number $c$ & MRR$\,\uparrow$ & Hits@c$\,\uparrow$ & F1$\,\uparrow$ \\
  \midrule 
    10 & 0.30 & 0.46 & 0.39 \\
    30 & 0.31 & 0.70 & 0.48 \\
    50 & 0.31 & 0.73 & 0.48 \\
  \bottomrule
 \end{tabular}
 \caption{Results of sensitivity analysis.}
 \label{tab:sa-c}
\end{table}

\section{Case Study of KBQA Results}
\label{sec:cs-kbqa}

\begin{figure*}[t]
\centering
\includegraphics[width=1\textwidth]{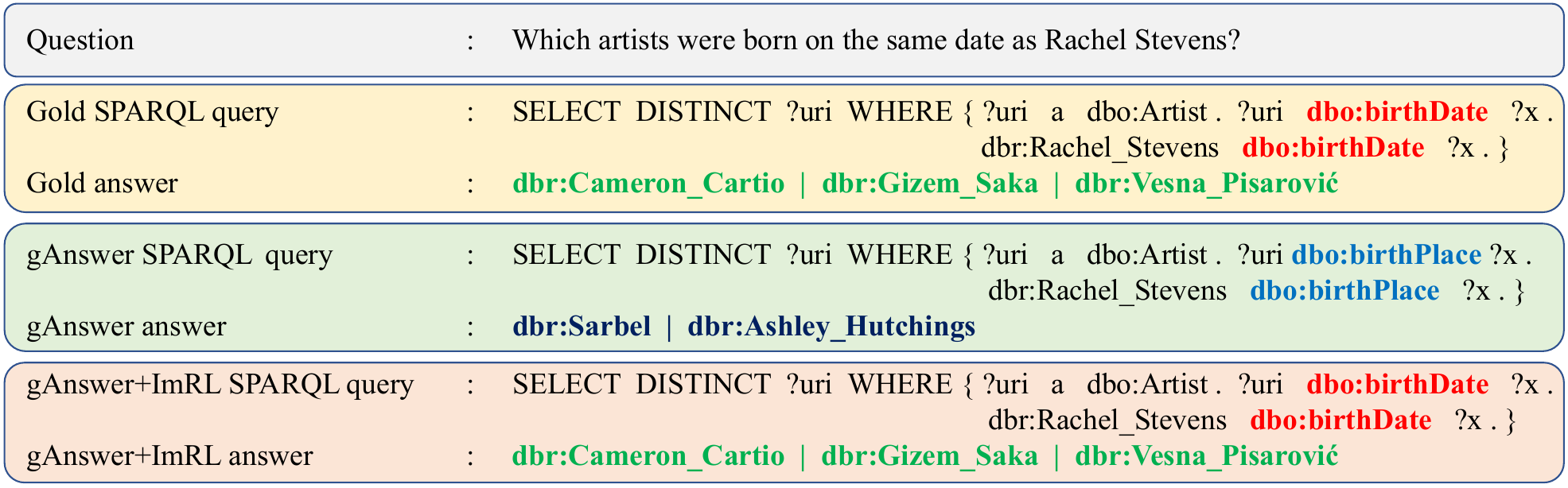}
\caption{Example 1 of KBQA with ImRL over DBpedia.}
\label{appendix:fig1}
\end{figure*}

\begin{figure*}[t]
\centering
\includegraphics[width=1\textwidth]{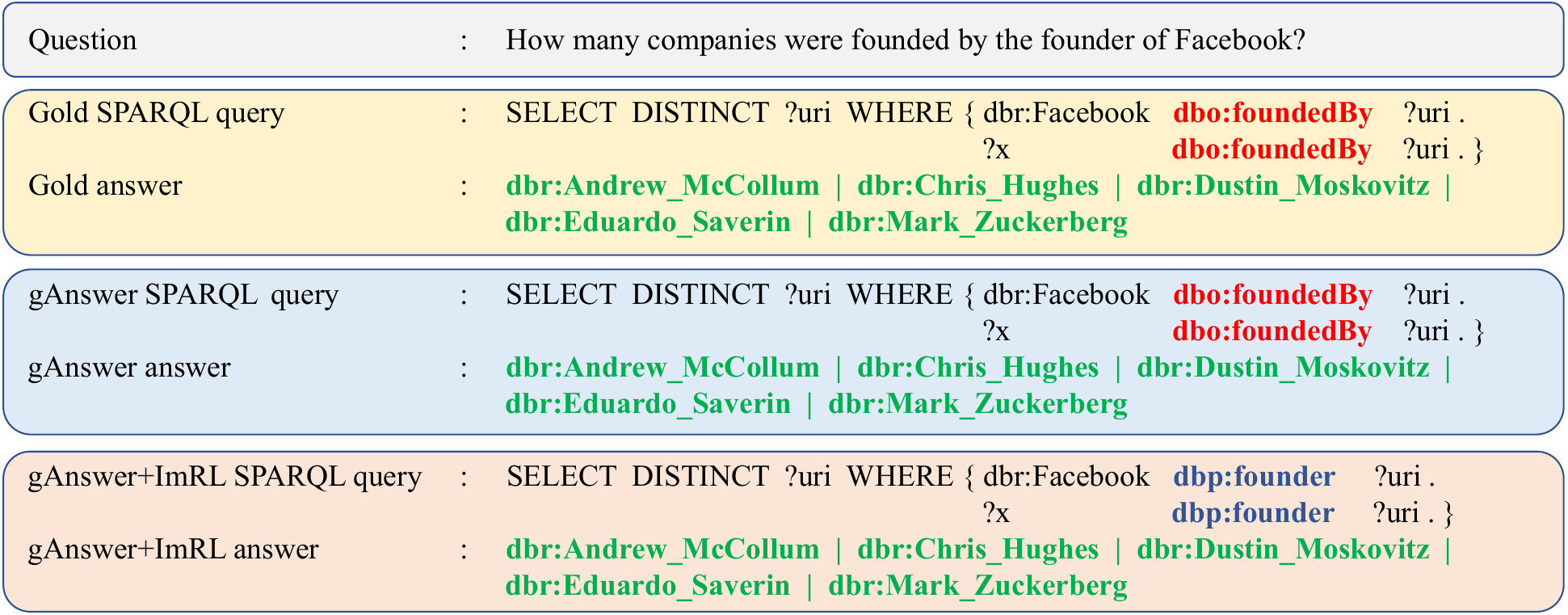}
\caption{Example 2 of KBQA with ImRL over DBpedia.}
\label{appendix:fig2}
\end{figure*}

By replacing the relation linking component in gAnswer (an open-source KBQA system) with ImRL, we conducted KBQA experiments over DBpedia. Compared with the original gAnswer, the overall F1-score increases by 3.2\%. Figure~\ref{appendix:fig1} and Figure~\ref{appendix:fig2} show two interesting cases. 
In Figure~\ref{appendix:fig1}, the relation linking module of gAnswer incorrectly links the relation in the question to \textit{dbo:birthPlace}, which leads to the wrong SPARQL query and the wrong answer. In contrast, ImRL can correctly identify the relation \textit{dbo:birthDate}, modify the SPARQL query and return the correct answer. The main reason is that ImRL can capture the context information ``date'' with the language model.
In Figure~\ref{appendix:fig2}, although ImRL does not recognize the same relation \textit{dbo:foundedBy} as the gold SPARQL query, it recognizes \textit{dbp:founder}, which is actually equivalent to \textit{dbo:foundedBy}. Due to the organizational structure of DBpedia, namespaces \textit{dbo} and \textit{dbp} have different representations of the same relations. However, with the help of KG representation learning, the embeddings of the two kinds of relations can be learned to have similar meanings. Thus, ImRL can capture similar relationships between \textit{dbo} and \textit{dbp} through the KG embedding module, which also shows that ImRL can leverage the structural information in the KG.

\section{Detailed Analysis of PathSQ}
\label{sec:da-pathsq}

\begin{figure*}[!h]
\centering
\includegraphics[width=1\textwidth]{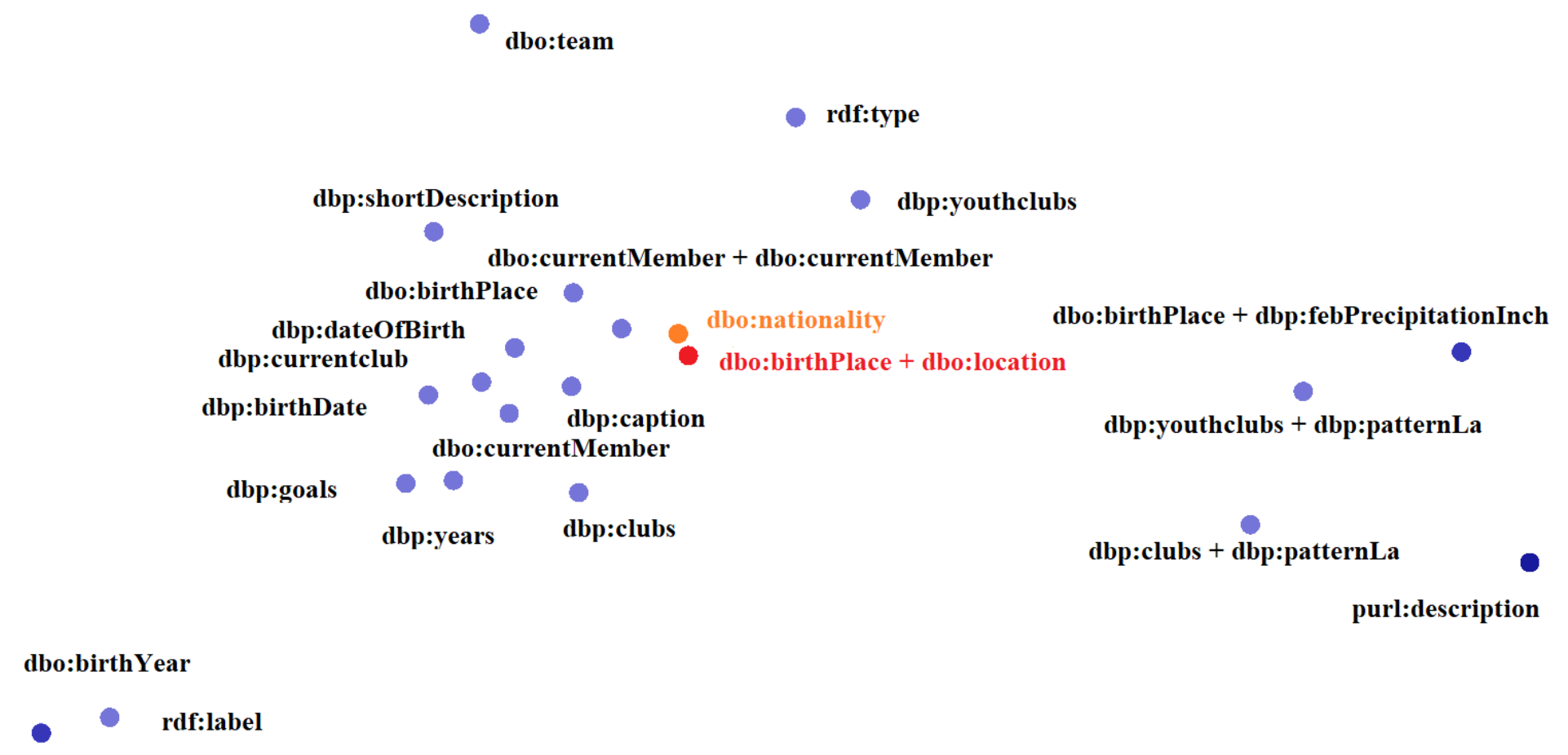}
\caption{Visualization of KG embeddings.}
\label{appendix:visual}
\end{figure*}

Since the questions in PathSQ all need to be linked through a multi-hop relation path in DBpedia, we further conduct a visualization experiment on PathSQ to validate whether KG embedding can eliminate the influence of incomplete KGs.
We use embeddings obtained through $R\_Encoder_{kg}$ as the input of Embedding Projector\footnote{\url{http://projector.tensorflow.org/}} to perform dimension reduction and obtain two-dimensional embeddings. 
Take the question ``which european nation is Rudi Hedman from?'' as an example, we enumerate the top-30 candidate relation paths of entity \textit{dbr:Rudi\_Hedman}. 
However, since entity \textit{dbr:Rudi\_Hedman} does not have relation \textit{dbo:nationality} in DBpedia, the gold relation path \textit{dbo:nationality} is missing in the candidate list.
As the phrase ``nation'' should be linked to \textit{dbo:nationality}, we compare these candidate relation paths with it.
In the KG embedding space, through the visualization (see Figure~\ref{appendix:visual}), we find that the embedding of the relation path \textit{dbo:birthPlace} $\rightarrow$ \textit{dbo:location} (color: red) is most similar to that of the relation \textit{dbo:nationality} (color: orange).
It shows that ImRL achieves implicit RL by expressing missing relation \textit{dbo:nationality} between the entities with relation path \textit{dbo:birthPlace} $\rightarrow$ \textit{dbo:location}.

\end{document}